\documentclass{article}

\usepackage[english]{babel}

\usepackage[preprint, nonatbib]{neurips_2024}

\usepackage{amsmath}
\usepackage{amsfonts}
\usepackage{graphicx} % For including images
\usepackage{algorithm, algpseudocode} % For algorithms
\usepackage{caption}  % for side-by-side table and figure in sec 3.1
\usepackage{authblk}

\usepackage[T1]{fontenc}
\usepackage[table, dvipsnames]{xcolor} % For extended color names
\usepackage[framemethod=TikZ]{mdframed} % For fancy framed boxes
\usepackage[colorlinks=true, linkcolor=violet, citecolor=YellowOrange, urlcolor=Aquamarine]{hyperref} % Pretty hyperlinks

\usepackage[authoryear, round]{natbib} 

\def\code#1{\texttt{#1}}

\usepackage{io}

\title{Iteration of Thought: Leveraging Inner Dialogue for Autonomous Large Language Model Reasoning}
\author[1]{Santosh Kumar Radha}
\author[1,2]{Yasamin Nouri Jelyani}
\author[1]{Ara Ghukasyan}
\author[1]{Oktay Goktas}
\affil[1]{Agnostiq Inc., 325 Front St W, Toronto, ON M5V 2Y1}
\affil[2]{ University of Toronto, 60 St George St, Toronto, Ontario, M5S 1A7, Canada}
\begin{document}
\maketitle

\begin{abstract}
Iterative human engagement is a common and effective means of leveraging the advanced language processing power of large language models (LLMs). Using well-structured prompts in a conversational manner, human users can effectively influence an LLM to develop more thoughtful and accurate responses. Motivated by this insight, we propose the Iteration of Thought (IoT) framework for enhancing LLM responses by  generating "thought"-provoking prompts \textit{vis a vis} an input query and the current iteration of an LLM's response. Unlike static or semi-static approaches, \textit{e.g.} Chain of Thought (CoT) or Tree of Thoughts (ToT), IoT adapts its reasoning path dynamically, based on evolving context, and without generating alternate explorative thoughts which are ultimately discarded. The three components of the IoT framework are (1) an Inner Dialogue Agent (IDA) responsible for generating instructive, context-specific prompts; (2) an LLM Agent (LLMA) that processes these prompts to refine its responses; and (3) an iterative prompting loop that implements a conversation between the former two components. We introduce two variants of our framework: Autonomous Iteration of Thought (AIoT), where an LLM decides when to stop iterating, and Guided Iteration of Thought (GIoT), which always forces a fixed number iterations. We investigate the performance of IoT across various datasets, spanning complex reasoning tasks from the GPQA dataset, explorative problem-solving in \textit{Game of 24}, puzzle solving in \textit{Mini Crosswords}, and multi-hop question answering from the HotpotQA dataset. Our results show that IoT represents a viable paradigm for autonomous response refinement in LLMs, showcasing significant improvements over CoT and thereby enabling more adaptive and efficient reasoning systems that minimize human intervention.\footnote{An installable implementation of the IoT framework can be found at \citep{multi_agent_llm_2024}}
\end{abstract}
\begin{figure}[h]
    \centering
    \includegraphics[width=\textwidth]{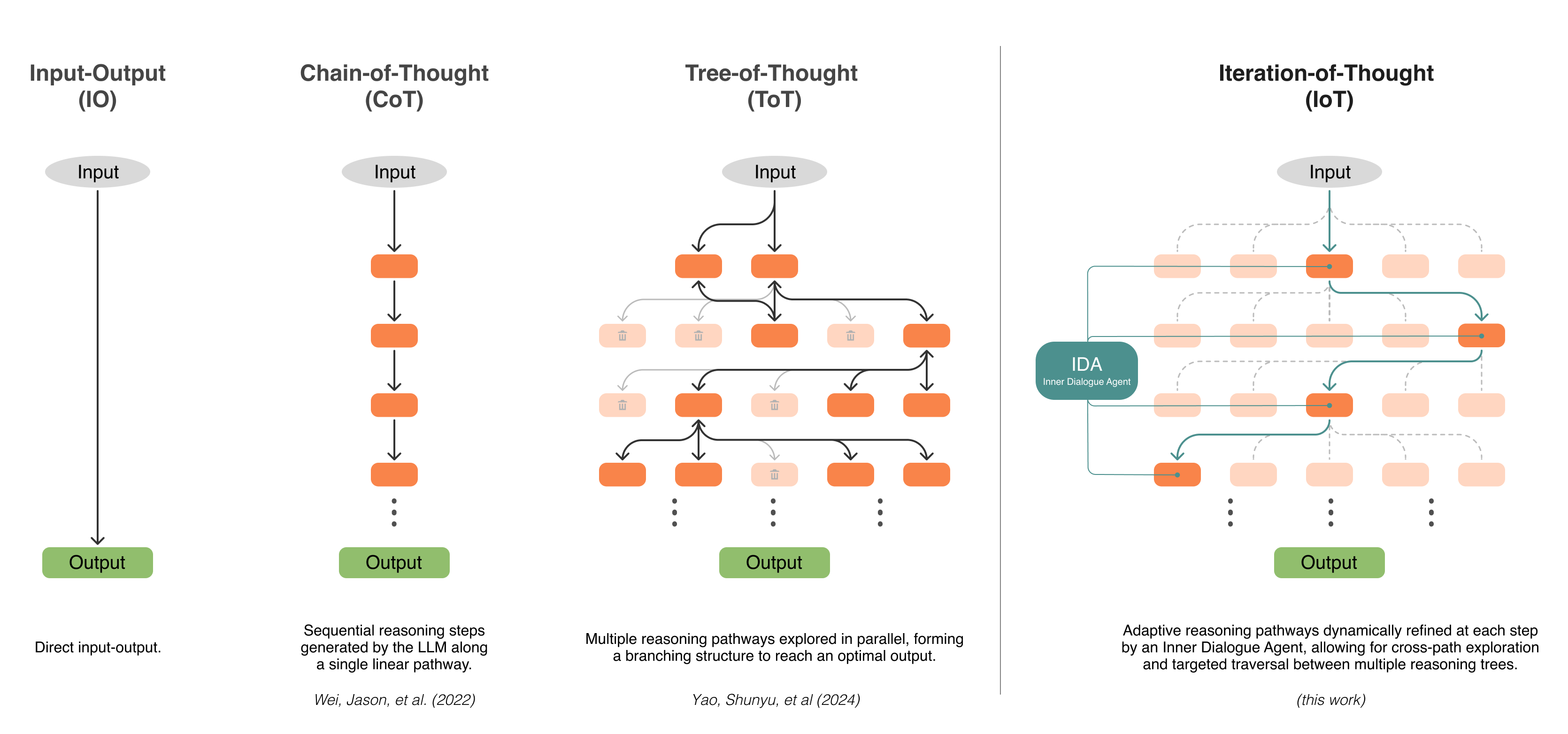}  % Replace with actual image path
    \caption{Illustration of different prompting strategies for enhancing LLM reasoning capabilities. The Input-Output (IO) method uses a direct input-output approach with no intermediate reasoning. Chain-of-Thought (CoT) \citep{wei2022chain} prompts introduce a single, linear reasoning path, while Tree-of-Thought (ToT) \citep{yao2024tree} methods expand this by exploring multiple reasoning paths in parallel. The proposed Iteration-of-Thought (IoT) (\textit{This work}) framework introduces an Inner Dialogue Agent (IDA) to dynamically adjust reasoning paths, enabling adaptive cross-path exploration to enhance response accuracy.}
    \label{fig:hero}
\end{figure}
\section{Introduction}

The development of Large Language Models (LLMs) like \code{GPT-3}, \code{PaLM} \citep{anil2023palm}, and their successors, including \code{GPT-4} \citep{openai2023gpt4}, \code{Gemini} \citep{team2023gemini}, \code{LLaMA} \citep{dubey2024llama}, and \code{Claude}, has revolutionized natural language processing. LLMs have empowered AI systems to perform a wide range of tasks with remarkable proficiency. In the context of human-LLM interaction, a critical observation from practical experience is that the quality of LLM responses tends to improve with repeated prompting and user feedback. Recent research demonstrated that naïve prompting can lead to calibration errors, while more sophisticated, iterative prompting strategies significantly improve both accuracy and reliability \citep{krishna2024understandingeffectsiterativeprompting}. These results suggest that, given context-appropriate sequences of inputs, LLMs can much more effectively leverage their internal knowledge base \citep{jiang2020can,petroni2019language,talmor2020olmpics,roberts2020much} to provide richer, more nuanced answers \citep{sloman1996empirical}.

A human user's interaction with in an LLM often proceeds as follows: the user poses a question to the LLM, receives an initial response, and, if the answer is incomplete or suboptimal, provides additional guidance to the LLM by reiterating contextual clues (\textit{e.g.} by reminding the LLM of its role, suggesting additional information to consider, or highlighting specific parts of the response that need refinement). This back-and-forth process helps narrow the focus of the LLM while reducing the research effort required from the user, since the LLM is responsible for the bulk of the reasoning and information retrieval.
% Recent studies have shown that LLMs are capable of generating not only relevant text, but also leveraging extensive knowledge \citep{jiang2020can,petroni2019language,talmor2020olmpics,roberts2020much} learned from diverse data sources, allowing them to provide answers that reflect a deep, encoded understanding of various domains.

We identify two predominant forms of human-LLM interaction. In the first form of interaction, the user simply guides an LLM through its own internal knowledge base. For example, consider a scenario where an LLM generates code that is syntactically incorrect due to a missing bracket. The user might prompt it to "verify the syntax," leading the LLM to correct the error in a subsequent response. In the second for of interaction, the user introduces new information to improve the LLM's response. For example, an LLM may be asked to provide up-to-date weather information for a specific city, but lacks access to real-time data. In this case, the user can supply this information (using a tool or API), then prompt the LLM to \textit{e.g.} recommend weather-appropriate clothing or destination to visit in that locale. All together, the first form an interaction leads the LLM to better utilize its \textit{internal} knowledge, whereas the second form of interaction involves augmenting the LLM's knowledge with new information.

The potential of iterative prompting to improve LLM responses is supported by research showing that prompt phrasing can significantly influence a model's performance in various settings \citep{brown2020language, opsahl2024optimizing}. Figure~\ref{fig:hero} illustrates the progression from simple Input-Output (IO) approaches to more advanced methods like Chain-of-Thought (CoT) \citep{wei2022chain} and Tree-of-Thought (ToT) \citep{yao2024tree}ThCoT introduces sequential reasoning steps along a single linear path, while ToT explores multiple reasoning pathways in parallel, forming a branching structure to optimize the output.

ese methods represent "reasoning frameworks"\citet{wei2022chain} that rely on static or semi-static prompts, which may struggle to adapt to the evolving context of each query and response, potentially limiting the quality of LLM responses. CoT prompting encourages LLMs to articulate its intermediate reasoning steps, which leads to better performance on complex tasks. Similarly, the related ToT approach (among other methods \citep{sahoo2024systematic}) reasons along multiple paths to consider a wider breadth of potential responses, most of which are generated then discarded, leading to better performance on more explorative tasks like solving puzzles or crosswords. Other frameworks like \textit{Self-Refine} \citep{madaan2024self} and \textit{Self-Verification} \citep{weng2022large} enable LLMs to iteratively critique and refine their outputs, but still rely on static or semi-static prompts. In a broader context, the value of pursuing improved reasoning with inference techniques, as opposed to extensive training, is underscored by more recent advancements such as OpenAI's new series of \code{o1} models \citep{openai2024o1}. These proprietary models are specifically designed to spend more time "thinking" through problems before responding, focusing on inference to solve complex tasks in science, coding, and math. Such developments highlight a broader shift in the AI community toward post-training enhancement of reasoning capabilities as a more scalable approach.

In this work, noting the lack of reasoning frameworks that strive to replicate the dynamic nature of human-LLM interactions, we propose IoT as an autonomous, iterative, and adaptive approach to LLM reasoning without human feedback.

\subsection{Iteration of thought (IoT)}

% Motivated by these insights, we propose the \textit{Iteration of Thought (IoT) framework} to emulate the dynamic interaction patterns observed in human-LLM exchanges. 
Unlike the aforementioned static and semi-static frameworks, IoT utilizes an Inner Dialogue Agent (IDA) to adjust and refine its reasoning path during each iteration. This enables adaptive exploration across different reasoning trees, fostering a more flexible and context-aware response generation process. A comparison to existing methods is shown schematically in Figure~\ref{fig:hero}.

The core IoT framework is composed of three main components. Further details are also provided in Section~\ref{sec:framework}.

\begin{itemize}

\item \textbf{Inner dialogue agent (IDA):} The IDA functions as a "guide" that dynamically generates context-sensitive prompts based on the original user query and the LLM’s previous response. The adjusted prompts servce to iteratively lead the LLM toward more refined and accurate answers. Mathematically, the IDA can be represented as a function \( C: \mathcal{Q} \times \mathcal{R} \times \mathcal{K}^\prime \rightarrow \mathcal{P} \), where \( \mathcal{Q} \) is the space of possible queries, \( \mathcal{R} \) is the space of potential LLM responses, and \( \mathcal{P} \) is the space of generated prompts. At each step, it takes the current query \( q \in \mathcal{Q} \) and the previous response \( r \in \mathcal{R} \) to generate a new prompt \( p \in \mathcal{P} \). This process makes prompt generation \textit{dynamic}, differentiating IoT from more rigid approaches like CoT and allowing it to adapt to an evolving context.

\item \textbf{LLM agent (LLMA):} The LLMA embodies the core reasoning capabilities of an LLM and processes the IDA's dynamically generated prompts. It uses an LLM's internal knowledge base \( K \) to refine its responses. Formally, we model the LLMA as a function \( L: \mathcal{Q} \times \mathcal{P} \times K \rightarrow \mathcal{R} \). The LLMA takes as input a query \( q \), prompt \( p \) and a knowledge base \( K \) then generates a refined response \( r \). The LLMA also identifies areas of uncertainty or gaps in its own reasoning, providing feedback for the IDA to adjust prompts accordingly. This interaction creates a closed-loop system that continuously improves the quality of answers without external inputs.

\item \textbf{Iterative prompting loop:} The iterative process in IoT involves a back-and-forth between the IDA and LLMA. At each iteration \( i \), the IDA generates a new prompt \( p_i = C(q, r_{i-1}) \) based on the original query \( q \) and the LLM's previous response \( r_{i-1} \). The LLMA then responds to $p_i$ with \( r_i = L(q, p_i, K) \). This loop continues until a satisfactory answer \( r^* \) is found or the arbitrary maximum iteration count is reached. This back-and-forth approach allows IoT to navigate complex reasoning paths to efficiently explore various potential solutions. Moreover, introducing distrinct LLMs for the IDA and LLMA respectively can allow each agent to function as an open system \citep{von1950theory} where internal knowledge is exchanged. In this scenario, the overall system behaves as a \textit{closed} system with a combined knowledge base, enhancing internal reasoning without external input.
\end{itemize}
% Our contributions can be summarized as follows: We introduce the IoT framework, which enables autonomous refinement of LLM responses by modeling the iterative interaction between an Inner Dialogue Agent and an LLM Agent. This dynamic interaction loop enables LLMs to iteratively assess and refine their answers without external input. 

In the sections that follow, we present a detailed analysis of our IoT framework, describe our experimental methodology, and discuss empirical results. We also demonstrate the framework's effectiveness with experimental results on the various datasets, where significant improvements are observed over existing reasoning methods.
%Furthermore, our framework opens up new possibilities for enhancing LLM performance through autonomous iterative refinement and lays the groundwork for future research on integrating multi-agent systems to further optimize each component's performance.

\section{Framework and implementation}\label{sec:framework}

In this work, we use two distinct variants of IoT: \textit{Autonomous Iteration of Thought} (AIoT) and \textit{Guided Iteration of Thought} (GIoT). In the AIoT variant, the LLMA itself decides when it has generated a satisfactory response. This decision is reflected in a Boolean output signal, \( \texttt{iteration\_stop} \). Termination following a positive signal usually leads to fewer iterations than the enforced maximum. This, in turn, leads to faster  evaluation with less exploration, but risks premature stops when facing more complex queries. Conversely, GIoT employs a more regimented strategy by mandating a fixed number of iterations. GIoT employs the opposite strategy, aiming for comprehensive exploration of reasoning paths to minimize premature convergence, at additional computational cost and with the risk of redundant or repetitive iterations.

We implemented the IoT framework, including both variants, as a Python library \citep{multi_agent_llm_2024}, using Pydantic \citep{pydantic_2024} to provide provide output schemas for raw responses from LLMs.

% The remainder of this section provides an in-depth explanation of the AIoT and GIoT variants of our framework.

% The IoT framework comprises two key components within an environment powered by one or more LLM engines: the Inner Dialogue Agent (IDA) and the LLM Agent (LLMA). Here, the IDA acts as a meta-agent, dynamically generating context-sensitive prompts based on the ongoing state of the conversation, aiming to steer the LLM toward better answers. A formal definition of either component is provided in the previous section. Recall that we represent the IDA as a function \( C: \mathcal{Q} \times \mathcal{R} \times \mathcal{K}^\prime \rightarrow \mathcal{P} \) and the LLMA as another function \( L: \mathcal{Q} \times \mathcal{P} \times K \rightarrow \mathcal{R} \).

\begin{figure}[h]
    \centering
    \includegraphics[width=\textwidth]{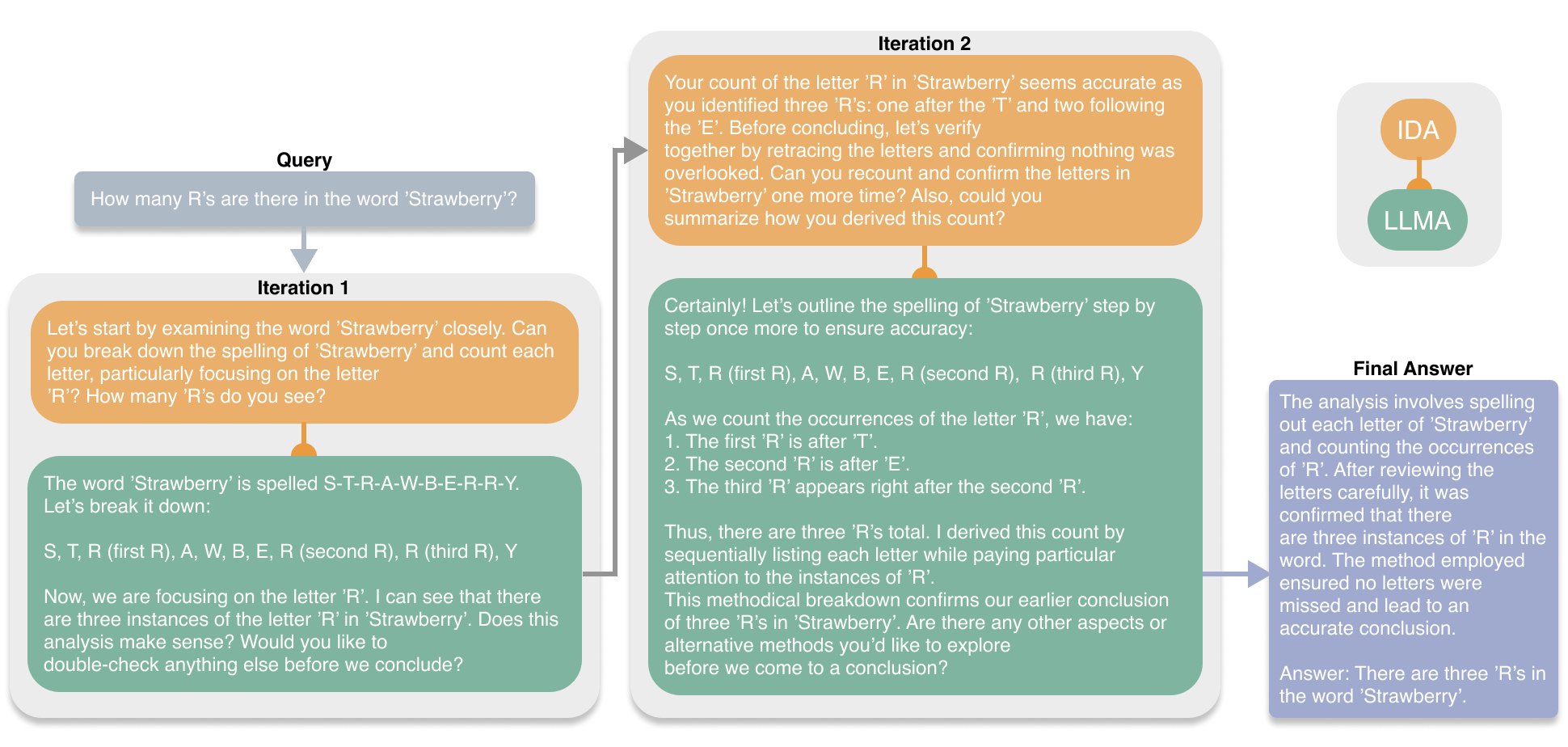}
    \caption{Schematic example of processing a sample query with the IoT framework. A simplistic question is asked for illustrative purposes. The guided IoT variant (GIoT) is utilized here, with the number of iterations set to 2. Each grey boxe contains an individual iteration of IoT, with the IDA shown in yellow and the LLMA in green.}
    \label{fig:giot_strawberry}
\end{figure}

% The IDA and LLMA interact in a loop. At each iteration \( i \), the IDA generates a new prompt \( p_i = C(q, r_{i-1}) \), to which the LLMA generates a response \( r_i = L(q, p_i, K) \). This process continues cyclically until a satisfactory response \( r^* \) is achieved or a maximum number of iterations \( T \) is reached.

% The interplay between the IDA and LLMA in the IoT framework mirrors the dynamic and adaptive nature of human-guided LLM interactions, where prompts are adjusted based on evolving context and outputs. This framework not only enhances the accuracy and depth of LLM responses but also reduces the need for extensive human intervention, paving the way for more autonomous and efficient reasoning systems.

\subsection{Autonomous iteration of thought (AIoT)}

% In this variant, the IDA guides the LLMA through iterative prompts, while adjusting the LLMA's focus based on both the query and the evolving conversation state. The LLM, configured as the LLMA, processes these prompts and provides refined responses. At each iteration \(i\), the IDA generates a new prompt \(p_i = C(q, r_{i-1})\) based on the initial query \(q\) and the previous response \(r_{i-1}\). The LLMA then generates a new response \(r_i = L(q, p_i, K)\), where \(K\) represents the LLM's internal knowledge base.
In AIoT, the LLM also makes a determination at each step on whether the answer it has generated is sufficient. This is represented by an output signal, \( \code{iteration\_stop} \), which, if set to \code{True}, indicates that the LLM believes its answer is final and complete.  The full AIoT process is shown in pseudocode Algorithm 1, below. A sample AIoT sequence is also provided in Appendix~\ref{appendix:experimental_details}.

\begin{algorithm}[H]
\caption{Autonomous Iteration of Thought (AIoT)}
\renewcommand{\algorithmicrequire}{\textbf{Input:}}
\renewcommand{\algorithmicreturn}{\textbf{Output:}}
\begin{algorithmic}[1]
\Require Query \( q \in \mathcal{Q} \), LLM configuration with IDA given by \( C: \mathcal{Q} \times \mathcal{R} \times \mathcal{K}^\prime \rightarrow \mathcal{P} \), LLMA given by \( L: \mathcal{Q} \times \mathcal{P} \times \mathcal{K} \rightarrow \mathcal{R} \), a maximum number of iterations \( T \in \mathbb{N}^+ \), and a stopping criterion given by \( \mathcal{F}: \mathcal{R} \times \mathcal{C} \rightarrow \{0,1\} \).
\State \( r_0 \gets L(q, \text{"Initial Prompt"}, \mathcal{K}) \) \Comment{Generate the initial response using LLMA}
\State \( i \gets 1 \) \Comment{Initialize the iteration counter}
\State \( \texttt{iteration\_stop} \gets \mathcal{F}(r_0, \mathcal{C}) \) \Comment{Evaluate stopping condition for the initial response}
\While{\( \neg \texttt{iteration\_stop} \land i \leq T \)} \Comment{Continue until stopping criteria or maximum iterations reached}
    \State \( p_i \gets C(q, r_{i-1}) \) \Comment{IDA generates a new prompt based on the query and the last response}
    \State \( r_i \gets L(q, p_i, \mathcal{K}) \) \Comment{LLMA generates a new response to the IDA prompt}
    \State \( \texttt{iteration\_stop} \gets \mathcal{F}(r_i, \mathcal{C}) \) \Comment{Evaluate stopping condition for the current response}
    \State \( i \gets i + 1 \) \Comment{Increment the iteration counter}
\EndWhile
\State \Return \( r_{i-1} \) \Comment{The last response that met stopping criteria or final response after \(T\) iterations}
\end{algorithmic}
\end{algorithm}

\subsection{Guided iteration of thought (GIoT)}

The guided variant of Iteration of Thought (GIoT) represents a more controlled iterative process. In GIoT, the iteration continues for a predefined number of steps \(N-1\), and only in the \(N\)-th iteration is the LLM allowed to decide if it has reached the final answer. Here, the IDA continues to generate new prompts \(p_i = C(q, r_{i-1})\) for the first \(N-1\) iterations without allowing the LLM to conclude early. In the final iteration, the LLMA is asked to provide a conclusive answer \(r^*\) based on the accumulated information from previous steps.

Like AIoT, GIoT ensures that the LLM thoroughly explores its internal knowledge space and refines its output to a greater extent. However, unlike AIoT, GIoT admits the cost of additional generations as a compromise to prevent premature conclusion. The full GIoT process is shown in pseudocode in Algorithm 2. A sample GIoT sequence is also provided in Figure~\ref{fig:giot_strawberry}.

\begin{algorithm}[H]
\caption{Guided Iteration of Thought (GIoT)}
\renewcommand{\algorithmicrequire}{\textbf{Input:}}
\renewcommand{\algorithmicreturn}{\textbf{Output:}}
\begin{algorithmic}[1]
\Require Query \( q \in \mathcal{Q} \), LLM configuration with IDA given by \( C: \mathcal{Q} \times \mathcal{R} \times \mathcal{K}^\prime \rightarrow \mathcal{P} \), LLMA given by \( L: \mathcal{Q} \times \mathcal{P} \times \mathcal{K} \rightarrow \mathcal{R} \), and a maximum number of iterations \( N \in \mathbb{N}^+ \).
\State \( r_0 \gets L(q, \text{"Initial Prompt"}, \mathcal{K}) \) \Comment{Generate the initial response using LLMA}
\For{\( i = 1, 2, \ldots, N-1 \)} \Comment{Iteratively refine the response for \(N-1\) iterations}
    \State \( p_i \gets C(q, r_{i-1}) \) \Comment{IDA generates a new prompt based on the query and previous response}
    \State \( r_i \gets L(q, p_i, \mathcal{K}) \) \Comment{LLMA generates a new response guided by the updated prompt}
\EndFor
\State \( p_N \gets C(q, r_{N-1}) \) \Comment{IDA generates the final prompt with explicit final instructions}
\State \( r^* \gets L(q, p_N, \mathcal{K}) \) \Comment{LLMA generates the final refined response} \\
\Return \( r^* \) \Comment{Return the final refined response after \(N\) iterations}
\end{algorithmic}
\end{algorithm}

To summarize, the choice of AIoT or GIoT defines the mode of iteration in the core IoT framework. Each variant allows the framework to approach the iterative refinement of LLM responses from different angles.

\section{Results}

To comprehensively evaluate the IoT framework, we conducted a series of experiments across various models, datasets, and reasoning strategies. Given the computational expense these evaluations, we selected specific model-dataset combinations to investigate performance and scalability under different conditions. This approach enables us to provide a targeted understanding of how reasoning capability and iteration strategy affect the overall quality of LLM responses. The following sections describe the experiments designed to explore these aspects, including their setups, objectives, and the insights derived from each.

\begin{figure}[h]
    \centering
    \begin{minipage}{0.45\textwidth}
        \centering
        \begin{tabular}{|l|c|c|}
        \hline
        \textbf{Method} & \textbf{Acc.} & \textbf{Improvement vs. IO (\%)} \\
        \hline
         IO & 0.405   & 0.00\%  \\
         CoT & 0.406  & 0.12\%  \\
         GIoT & 0.416 & 2.62\%  \\
         AIoT & \textbf{0.463} &  \textbf{14.11\%} \\
        \hline
        \end{tabular}
        \captionof{table}{Comparison of accuracies (and relative improvements) for different methods on GPQA Diamond Dataset.}
        \label{tab:gpqa_comparison}
    \end{minipage}%
    \hfill
    \begin{minipage}{0.45\textwidth}
        \centering
        \includegraphics[width=\textwidth]{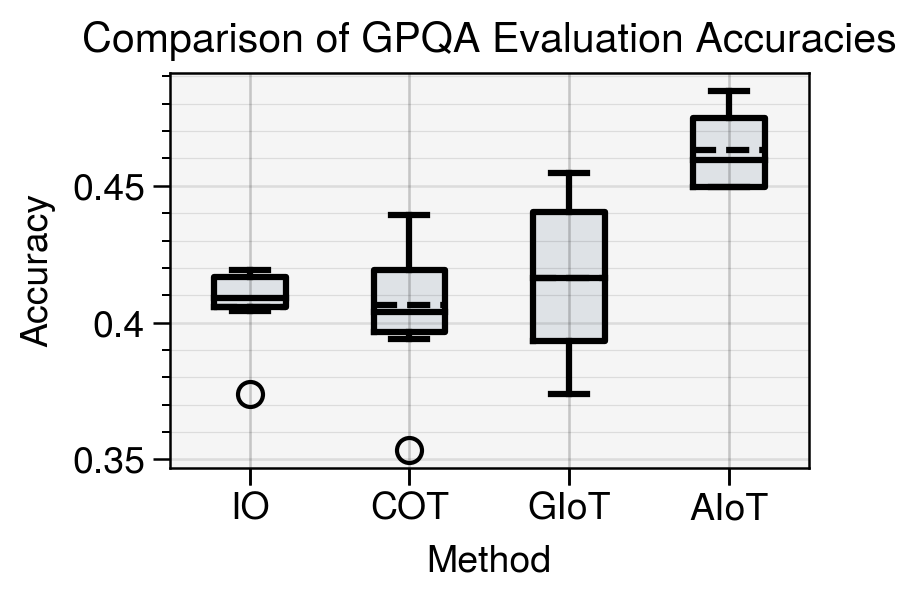}
        \caption{Comparison of GPQA evaluation accuracies for different methods.}
        \label{fig:gpqa_results}
    \end{minipage}
\end{figure}

\subsection{Assessing IoT on the GPQA questionnaire}
% \san{We can move some of this to appendix and make the initial discussion about models etc.. smaller to focus on result and discussion.}

% \ara{changing this to remove model size component for now}

In this experiment, we quantify IoT's ability to accurately answer questions from the GPQA Diamond dataset \citep{rein2023gpqa}. These questions are known to require deep reasoning and comprehensive internal knowledge, with even the highly capable LLMs yielding overall scores under 50\% \citep{dubey2024llama}. 

We compare AIoT/GIoT with CoT on \code{GPT-4o mini}, a proprietary model \citep{openai2023gpt4}. CoT is a widely used reasoning strategy that involves guiding the model through a step-by-step thinking process. By comparing CoT with AIoT/GIoT, we aim to understand the potential improvements resulting from iterating \textit{dynamically} rather than following predefined steps.

The results of these experiments are presented in Table~\ref{tab:gpqa_comparison} andFigure~\ref{fig:gpqa_results}ere it, is clear that conventional CoT performs about on par with the baseline IO approach, indicating that rigid step-by-step reasoning may not be effective for GPQA. Meanwhile, GIoT performs significantly better than IO and CoT with a modest 2.62\% higher accuracy (on average). However, GIoT also exhibits larger variance than IO and CoT in its distribution of accuracy scores. One interpretation of this result is that forced iterations can lead to divergence due to hallucination \citep{huang2023survey} in cases where a correct and complete thought pattern is established well before the mandated number of iterations have been performed. AIoT, on the other hand, is more effective at avoiding this issue.

 emerges as the most effective strategy overall, with a  14.11\% improvement in average accuracy over the IO baseline and the lowest variance among all methods tested. Lower variance in AIoT's accuracy scores implies more consistent performance across different types of questions. Together with a higher average score, this superior result is attributed to AIoT's dynamic autonomous, context-aware termination of iterations, which prevents unproductive or counterproductive exploration of the response space. Notably, our analysis shows that AIoT completes approximately 60\% of tasks within a single iteration and approximately 90\% within two iterations. This reflects AIoT's efficiency in navigating the reasoning space without over-iteration. We therefore infer that AIoT's advantage, wherever applicable, is avoiding the pitfalls of both under- (as seen in IO and CoT) and over-exploration (a risk associated with GIoT).

\subsection{Assessing IoT on explorative problem-solving tasks}
To evaluate the effectiveness of our IoT (Iterative of Thought) framework against the state-of-the-art, we conduct a comparative analysis using the \textit{Game of 24} and \textit{Mini Crosswords} tasks. These games, featured prominently in the ToT genesis paper by \citet{yao2024tree}, are easy to understand, challenging to solve, but easy to verify. ToT is well-suited for problems that benefit from a wide variety of exploratory reasoning paths, owing to its systematic search strategy that traverses many possible solution graphs to find the optimal answer. Our motivation for this experiment is to assess whether our IoT method can effectively iterate towards optimal solutions without generating a multitude of alternate, discarded responses. With this in mind, our goal in this experiment is to compare the relative advantage of our IoT framework compared to CoT, recognizing the inherent advantages of ToT, at least in terms its overall solution ability, in contexts benefiting from a broader exploratory approach.

The \textit{Game of 24} involves generating an arithmetic expression using four given numbers and basic operations and brackets $\{+, -, \times, \div,\left(,\right)\}$ to arrive at the number 24. This task requires not only computational ability but also strategic reasoning to explore different combinations of operations. The dataset for this task, as used in the ToT study, consists of various instances where the challenge lies in finding the most efficient path to the solution amidst multiple possibilities. Similarly, the \textit{Mini Crosswords} task involves solving 5x5 crossword grids based on a set of clues. Solving these grids requires lexical reasoning and pattern recognition, as well as the ability to generate coherent word sequences that fit both vertical and horizontal constraints. The complexity of the \textit{Mini Crosswords} task also stems from the need to test the compatibility of multiple potential word fits, refining choices based on feedback and constraints. Both datasets are therefore valuable for assessing a model's ability to try out various solutions within a reasonable search space.

% The iterative nature of the IoT method allows for dynamic exploration while maintaining the flexibility to focus on the most promising paths, rather than exhaustively exploring all possibilities like ToT. This experiment will help us determine how well IoT adapts to tasks that benefit from multiple iterations and potential solutions, providing insight into the model's capability to balance exploration with targeted refinement in scenarios where the optimal reasoning path is not immediately clear. Given these conditions, our focus is on comparing the relative gains from CoT to IoT within our framework and examining how these gains align with or differ from those reported for ToT, recognizing the inherent advantages of ToT in broader exploratory contexts.

\begin{figure}[h]
    \centering
    \includegraphics[width=\textwidth]{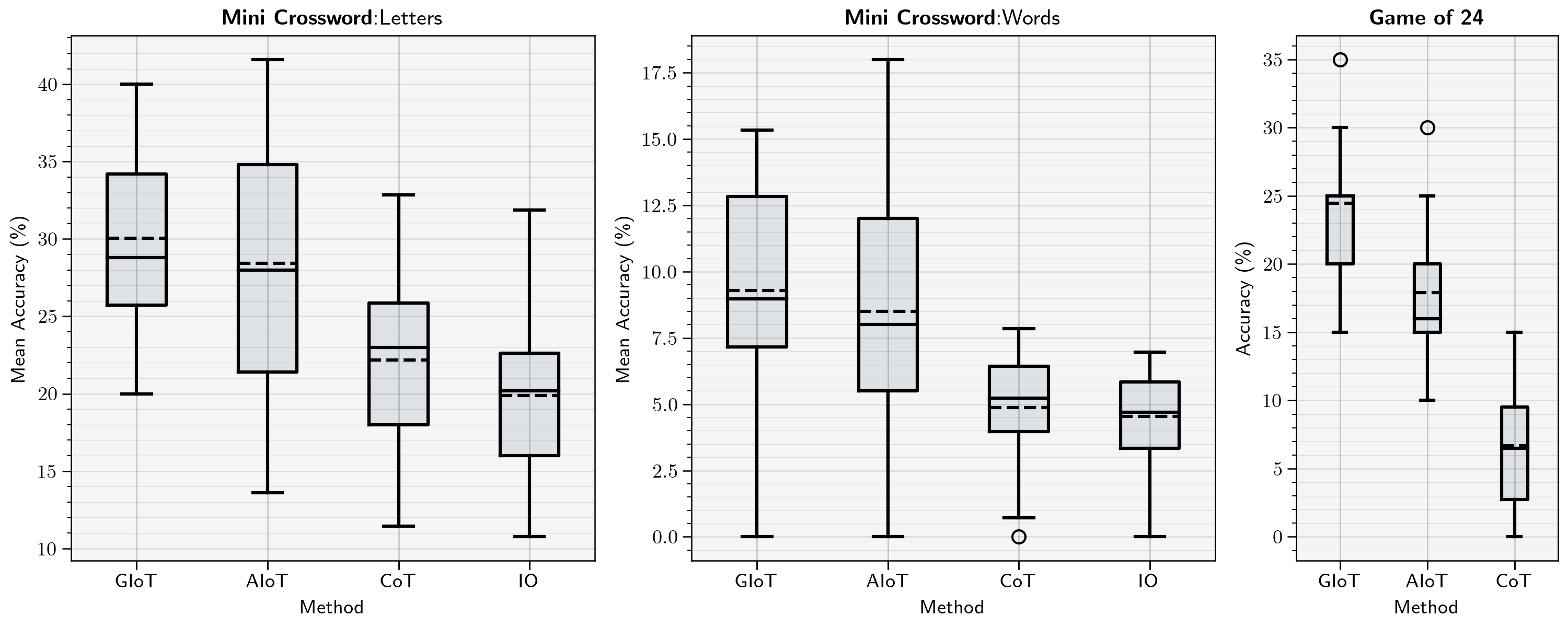}  % Replace with actual image path
    \caption{Performance comparison across different methods (GIoT, AIoT, CoT, IO) on Mini Crossword: Letters, Mini Crossword: Words, and Game of 24 tasks. Box plots represent the distribution of mean accuracy percentages across different trials.}
    \label{fig:performance_comparison}
\end{figure}

Results for these tasks are visualized in Figure~\ref{fig:performance_comparison}, which reveals distinct performance differences between the various methods. Notably, GIoT on average outperforms the AIoT, CoT, and IO methods across both tasks. This result is consistent with the understanding that GIoT is a more exploratory alternative to AIoT. By compelling the model to explore multiple reasoning paths, GIoT enhances the likelihood of arriving at a correct answer, aligning with the ToT approach in terms of beneficial brute-force exploration.

Regarding the \textit{Mini Crosswords} task, the original ToT study (which used \code{GPT-4}) demonstrated substantial improvements over CoT, with success rate increases of 92.1\% for letters and 284.6\% for words \citep{yao2024tree}. In comparison, our experiments using the less capable \code{GPT-4o mini} model show that GIoT achieves a success rate of 35.5\% for letters and a 90.6\% success rate for words, as compared to CoT. Meanwhile, AIoT shows gains of 28.3\% and 74.5\%, respectively. Although these differences are smaller than those reported for ToT, they should be considered in context with the limitions of \code{GPT-4o mini} versus \code{GPT-4}. It is also important to note that the superior performance of ToT in this task is primarily due to its capacity to explore a broader range of answers, potentially admitting a higher computational cost than GIoT.

The higher variance observed in the IoT results, particularly in the \textit{Mini Crosswords} task, suggests a more diverse albeit not always productive exploration of solutions compared to CoT and IO. While this diversity can be advantageous in some scenarios, it may lead to sub-optimal convergence in more constrained problem spaces.

A similar pattern of performance differences emerges in the \textit{Game of 24} task. Here, the ToT framework showed a dramatic improvement, with success rates increasing from 4.0\% with CoT to 74\% with ToT (at a breadth of 5), marking a relative improvement of 1750\% \citep{yao2024tree}. In comparison, our GIoT method achieves a notable 266.4\% improvement over CoT, while AIoT shows a 168.4\% increase. These results reflect the effectiveness of our iterative refinement approach in arithmetic problem-solving scenarios, even though a performance gap remains compared to ToT. The structured, multi-step reasoning of GIoT ensures a more thorough exploration of the solution space, which aligns with the exploratory nature of ToT but operates within the constraints of our closed-system approach. A key distinction between our method and ToT is the feedback mechanism: while ToT benefits from its ability to explore more extensive solution spaces or receive external correctness checks, our methods, especially AIoT, can lead to cases where incorrect answers are confidently selected. Integrating external validation tools or feedback could therefore significantly enhance IoT's performance on this and similar tasks.

%---------

\subsection{Assessing IoT on multi-context reasoning and retrieval tasks}
In our final experiment, we evaluate IoT on the HotpotQA-Hard dataset, a challenging benchmark for multi-hop question answering that demands sophisticated aggregate reasoning. Unlike simpler tasks that require straightforward information retrieval, HotpotQA involves complex information synthesis across multiple documents, requiring models to shift focus between various contexts to build a coherent answer. This necessitates bridging implicit information gaps, resolving ambiguities, and integrating scattered evidence.

Answering a HotpotQA question often involves several interconnected steps where initial findings must be used to guide further evidence retrieval. This process mirrors the key strengths of IoT: its ability to adaptively explore different reasoning paths, dynamically integrate context, and iteratively refine conclusions. The IoT's IDA plays a pivotal role here by guiding the LLMA to revisit and adjust its focus based on intermediate outputs, promoting more comprehensive exploration of the problem space. Such a mechanism is crucial for HotpotQA tasks, where the model must constantly re-evaluate earlier conclusions in light of newly synthesized information, ultimately leading to a more robust and accurate final answer.

For this experiment, again using \code{GPT-4o mini} as our engine, we benchmark the performance of AIoT against CoT using on three evaluation metrics: Exact Match (EM), F1 score, and ROUGE-L score. These metrics capture different facets of multi-hop QA performance: EM measures the proportion of exact matches with the ground truth, providing a stringent gauge of model accuracy; the F1 score balances precision and recall, capturing partial correctness; and ROUGE-L evaluates the longest common sub-sequence between generated and reference answers, highlighting semantic coherence.

The dynamic nature of (A)IoT allows it to autonomously adapt the depth of reasoning based on the complexity of the query, facilitating a flexible exploration of reasoning paths that CoT's static, step-by-step approach may lack. This flexibility enables the IoT framework to better handle the inherent ambiguities of HotpotQA, such as resolving conflicts or disambiguating entities across contexts. This can also serve as a self-correcting mechanism, helping to recognize gaps or errors in reasoning early on and prompting further exploration in subsequent iterations.

\begin{figure}[h]
    \centering
    \includegraphics[width=\textwidth]{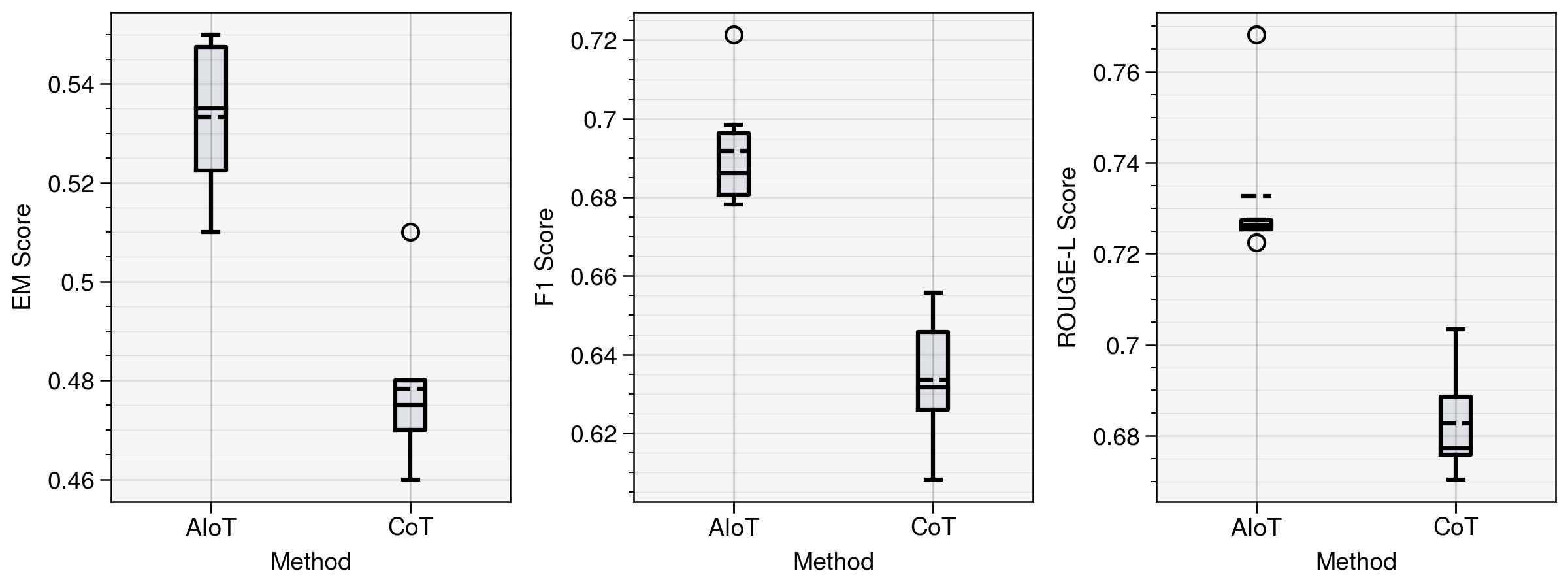}  
    \caption{Performance comparison between AIoT and CoT on HotPotQA-Hard dataset}
    \label{fig:hotpot}
\end{figure}

Our results on the HotpotQA-Hard dataset reveal a clear advantage for AIoT over CoT. As shown in Figure~\ref{fig:hotpot}, AIoT achieves an Exact Match (EM) score of 0.53, an F1 score of 0.699, and a ROUGE-L score of 0.72, significantly outperforming CoT on almost every instance of the task. These metrics demonstrate the effectiveness of AIoT in managing the complexities inherent in multi-hop question answering tasks, where models must dynamically integrate and synthesize information from various documents.

The observed performance gains are consistent with the core principles underlying the IoT framework. Using an Inner Dialogue Agent that steers the reasoning path, AIoT can better navigate ambiguities and implicit connections between different pieces of information. Improvements in the F1 score are indicative of AIoT’s ability to partially correct initial errors and refine its answers over multiple iterations. Similarly, the higher ROUGE-L score reflects AIoT's capacity to generate answers that are not only factually correct but also maintain semantic alignment with the ground truth. These results validate our hypothesis that iterative, adaptive reasoning is essential for tasks requiring complex information synthesis across disjoint contexts.

To contextualize our findings, we compare our AIoT approach with the \textit{AgentLite} framework introduced by \citet{liu2024agentlite}. \textit{AgentLite} is built on a novel hierarchical multi-agent orchestration technique that supports structured multi-agent systems, where a manager agent coordinates a set of team agents, each handling different aspects of reasoning. In their experiments on the HotpotQA dataset, \citet{liu2024agentlite} utilized \textit{AgentLite} to implement agents capable of multi-hop reasoning across multiple documents. The action space for these agents was designed with three primary members: \textit{WikipediaSearch}, \textit{Think}, and \textit{Finish}. Various models were tested within this framework, including \code{GPT-4}, a generally more knowledgeable model \citep{OpenAI2024eval} than \code{GPT-4o mini} utilized in our experiments.

\begin{table}[h]
\centering
\begin{tabular}{|l|l|c|c|}
\hline
\textbf{Method} & \textbf{Model} & \textbf{F1 Score} & \textbf{Exact Match (EM) Score} \\ \hline
Multi-Agent \textit{(AgentLite)} & \code{GPT-4-0613} & 0.527 & 0.38 \\ 
Multi-Agent \textit{(AgentLite)} & \code{GPT-4-32k-0613} & 0.520 & 0.37 \\ 
AIoT \textit{(Ours)} & \code{GPT-4o mini} & \textbf{0.699} & \textbf{0.53} \\ \hline
\end{tabular}
\caption{Performance Comparison on HotpotQA-Hard Dataset between AIoT Framework and AgentLite Benchmarks.}
\label{tab:comparison}
\end{table}

Comparing results from our AIoT approach to \textit{AgentLite} (Table~\ref{tab:comparison}) shows that AIoT achieves higher F1 and EM scores across the board. The AIoT framework’s F1 score of 0.699 and EM score of 0.53 surpass the results of even the most potent models used in the \textit{AgentLite} experiments, such as \code{GPT-4-0613} and \code{GPT-4-32k-0613}. This suggests that while \textit{AgentLite} offers a robust approach to structured reasoning, it may lack the adaptability and refinement capabilities that AIoT provides. By focusing on an autonomous, self-guided iteration process, AIoT effectively revisits and recalibrates its reasoning, allowing for deeper context integration and a more comprehensive exploration of the problem space. This comparison also validates the advantages of our approach in leveraging dynamic reasoning to outperform more static agentic frameworks in a multi-hop QA scenario.

To further illustrate the effectiveness of the IoT framework, we note that it also outperforms recent methods for multi-hop reasoning on HotpotQA, such as those described by \citet{wang2024adapting}, \citet{wang2024knowledge}, and \citet{jiapeng2024tree}. Although these studies demonstrate improvements over the CoT approach, the increase in F1 and EM scores achieved by IoT is larger than those reported in the aforementioned works. While not all these studies utilize \code{GPT-4o mini}, which makes direct comparisons less straightforward, it remains evident that the jump in accuracy from CoT to our IoT framework is much more pronounced.

\section{Strengths and weaknesses of IoT}

One qualitative benefit of IoT is its inherent conceptual transparency and explainability. Like CoT and similar methods, IoT provides a clear trace of its reasoning process through a sequence of evolving outputs. However, unlike other "multi-thought" methods, IoT's sequence also includes explicit guidance generated by the IDA. This means each step is accompanied by a rationale that the underlying LLM treats equivalently to prompts from a human user. As a result, \textit{post hoc} analysis of IoT's output sequences (example in Appendix~\ref{appendix:experimental_details}) can reveal the model's capacity to self-correct when provided with course-adjusting instructions. In addition to enhancing the model's explainability, this insight can inform more efficient interactions with LLMs in general.

It is important to note that the IoT framework is not inherently orthogonal to CoT nor \textit{Self-Consistent} CoT \citep{wang2022self}. One could in principle combine IoT with CoT to create a hybrid method, IoT$\circ$CoT, where both the Inner Dialogue Agent (IDA) and LLM Agent (LLMA) use CoT-based reasoning. Such combinations could amplify the benefits of structured reasoning while retaining the flexibility of iterative refinement. Additionally, while our experiments used the same base LLM for both IDA and LLMA, these agents can be made distinct to leverage different models or architectures, changing the total base knowledge of the system to be $K \otimes K^\prime$ \citep{creswell2022faithful}.

Owing largely to the versatility of LLMs, agentic LLM-based frameworks are not difficult to expand and compose. Recent work has suggested that larger ensembles of agents can lead to better reasoning performance \citep{li2024more}, with the rate of improvement diminishing beyond 10-15 agents. A natural progression of IoT could therefore be an expansion of the IDA into a \textit{meta-agent} consisting of specialized sub-agents, which may or may not be dynamically defined on a per-query basis. Taking the knowledge base of the IDA to be $K^\prime = \bigotimes_j^M K^j$, the size of the IDA ensemble, $M \in \mathbb{N}^+$, becomes an arbitrary parameter indicating the number of distinct LLMs behind the IDA's constituent sub-agents. IoT as introduced in this work (with $K = K^\prime$) represents the "smallest" member of this generalized family and still suffices to deliver powerful reasoning capabilities. Based on \cite{li2024more}, increasing the ensemble size $M$ could be expected to improve reasoning performance in IoT, though at cost of additional complexity and a larger hardware budget to power a multitude of distinct LLMs. $M>1$ also introduces the potentially challenging task of ranking sub-agent outputs or resolving conflicts in their guidance. Regarding complexity, using larger LLMs in a smaller ensemble may be a viable alternative for increasing the size of $K^\prime$ without increasing $M$.

IoT's autonomous iteration also offers significant advantages in situations where human intervention is impractical or impossible --- such that systems are constrained to function independently. Human oversight is difficult to achieve in contexts that demand rapid and continuous decision-making, for example. Here, IoT's autonomous reasoning capabilities can be a valuable asset. Moreover, the thought sequences generated by IoT (see Figure~\ref{fig:giot_strawberry} and Appendix~\ref{appendix:experimental_details}) could serve as a valuable resource for fine-tuning existing models, potentially enhancing their reasoning capabilities. This dual benefit of autonomy and improved model training makes IoT a powerful tool in building more robust, self-sufficient systems.

Regarding the two variants of IoT, our results demonstrate that while AIoT provides an efficient approach with autonomous decisions to stop iterating, it also often misjudges the completeness of its responses, leading to premature convergence. This limitation could be addressed by incorporating feedback agents \citep{chen2023teaching}, using techniques like \textit{maieutic prompting} \citep{jung2022maieutic}, or even allowing for human intervention or external knowledge checks. This would create a semi-autonomous framework that balances efficiency with robustness \citep{wu2022survey}. On the other hand, GIoT forces a fixed number of iterations, which can improve performance in multi-step reasoning tasks, but may also increase the risk of hallucination if the model confidently drifts into incorrect reasoning. Appropriate techniques to reduce hallucination could further refine GIoT's utility in complex tasks \citep{tonmoy2024comprehensive}.

\section{Conclusion and future work}

In this work, we introduced the Iteration of Thought (IoT) framework, in which an Inner Dialogue Agent (IDA) iteratively converses with an LLM Agent (LLMA) to perform various complex reasoning tasks like solving puzzles (\textit{Game of 24}, \textit{Mini Crosswords}) and answering difficult questionnaires (GPQA, HotpotQA). We employed two variants of this framework in our experiments, qualified as "autonomous" (AIoT) and "guided" (GIoT) respectively, to compare iteration-terminating mechanisms across these tasks. GIoT, the variant that always performs a fixed number of iterations, was seen to perform better than AIoT, the variant that self-determines termination, in \textit{Game of 24}. On the other hand, AIoT had superior performance on GPQA. Both variants performed similarly on \textit{Mini Crosswords} and always performed better than the well-known Chain of Thought (CoT) framework, wherever compared. We also compared our IoT framework against the hierarchical \textit{AgentLite} framework on the multi-context HotpotQA task, finding improvements of approximately a 35\% in the F1 score and 44\% in the EM score over \textit{AgentLite}. All together, our results demonstrate that IoT can succesfully introduce productive dynamism into low-complexity agentic frameworks. 

Determining the scale and diversity of the IDA's knowledge base represents a promising direction for future work aiming to maximize the real-world utility of IoT. In pursuit of strictly framework-to-framework comparisons, we used only off-the-shelf, general-purpose LLMs in all our experiments to establish IoT. Moving forward, specialized language models like fine-tuned LLMs or LLMs equipped with additional tools and/or data sources could yield further performance gains, whether by increasing the effective knowledge base or directly addressing challenges like hallucination and the premature termination of iterations. 
% Implementation of AIoT and GIoT methods and experimental result data are updated in the GitHub repository \citep{multi_agent_llm_2024}.

\bibliographystyle{plainnat}
\bibliography{bib}

\begin{thebibliography}{35}
\providecommand{\natexlab}[1]{#1}
\providecommand{\url}[1]{\texttt{#1}}
\expandafter\ifx\csname urlstyle\endcsname\relax
  \providecommand{\doi}[1]{doi: #1}\else
  \providecommand{\doi}{doi: \begingroup \urlstyle{rm}\Url}\fi

\bibitem[AgnostiqHQ(2024)]{multi_agent_llm_2024}
Github AgnostiqHQ.
\newblock multi-agent-llm-0.1.2.
\newblock \url{https://github.com/AgnostiqHQ/multi-agent-llm}, 2024.
\newblock URL \url{https://github.com/AgnostiqHQ/multi-agent-llm}.
\newblock Implementation of the AIOT and GIOT methods for multi-agent LLM architectures.

\bibitem[Anil et~al.(2023)Anil, Dai, Firat, Johnson, Lepikhin, Passos, Shakeri, Taropa, Bailey, Chen, et~al.]{anil2023palm}
Rohan Anil, Andrew~M Dai, Orhan Firat, Melvin Johnson, Dmitry Lepikhin, Alexandre Passos, Siamak Shakeri, Emanuel Taropa, Paige Bailey, Zhifeng Chen, et~al.
\newblock Palm 2 technical report.
\newblock \emph{arXiv preprint arXiv:2305.10403}, 2023.

\bibitem[Brown(2020)]{brown2020language}
Tom~B Brown.
\newblock Language models are few-shot learners.
\newblock \emph{arXiv preprint arXiv:2005.14165}, 2020.

\bibitem[Chen et~al.(2023)Chen, Lin, Sch{\"a}rli, and Zhou]{chen2023teaching}
Xinyun Chen, Maxwell Lin, Nathanael Sch{\"a}rli, and Denny Zhou.
\newblock Teaching large language models to self-debug.
\newblock \emph{arXiv preprint arXiv:2304.05128}, 2023.

\bibitem[Creswell and Shanahan(2022)]{creswell2022faithful}
Antonia Creswell and Murray Shanahan.
\newblock Faithful reasoning using large language models.
\newblock \emph{arXiv preprint arXiv:2208.14271}, 2022.

\bibitem[Dubey et~al.(2024)Dubey, Jauhri, Pandey, Kadian, Al-Dahle, Letman, Mathur, Schelten, Yang, Fan, et~al.]{dubey2024llama}
Abhimanyu Dubey, Abhinav Jauhri, Abhinav Pandey, Abhishek Kadian, Ahmad Al-Dahle, Aiesha Letman, Akhil Mathur, Alan Schelten, Amy Yang, Angela Fan, et~al.
\newblock The llama 3 herd of models.
\newblock \emph{arXiv preprint arXiv:2407.21783}, 2024.

\bibitem[Huang et~al.(2023)Huang, Yu, Ma, Zhong, Feng, Wang, Chen, Peng, Feng, Qin, et~al.]{huang2023survey}
Lei Huang, Weijiang Yu, Weitao Ma, Weihong Zhong, Zhangyin Feng, Haotian Wang, Qianglong Chen, Weihua Peng, Xiaocheng Feng, Bing Qin, et~al.
\newblock A survey on hallucination in large language models: Principles, taxonomy, challenges, and open questions.
\newblock \emph{arXiv preprint arXiv:2311.05232}, 2023.

\bibitem[Jiang et~al.(2020)Jiang, Xu, Araki, and Neubig]{jiang2020can}
Zhengbao Jiang, Frank~F Xu, Jun Araki, and Graham Neubig.
\newblock How can we know what language models know?
\newblock \emph{Transactions of the Association for Computational Linguistics}, 8:\penalty0 423--438, 2020.

\bibitem[Jiapeng et~al.(2024)Jiapeng, Runze, Yabo, Tong, Mingling, and Xiang]{jiapeng2024tree}
Li~Jiapeng, Liu Runze, Li~Yabo, Zhou Tong, Li~Mingling, and Chen Xiang.
\newblock Tree of reviews: A tree-based dynamic iterative retrieval framework for multi-hop question answering.
\newblock \emph{arXiv preprint arXiv:2404.14464}, 2024.

\bibitem[Jung et~al.(2022)Jung, Qin, Welleck, Brahman, Bhagavatula, Bras, and Choi]{jung2022maieutic}
Jaehun Jung, Lianhui Qin, Sean Welleck, Faeze Brahman, Chandra Bhagavatula, Ronan~Le Bras, and Yejin Choi.
\newblock Maieutic prompting: Logically consistent reasoning with recursive explanations.
\newblock \emph{arXiv preprint arXiv:2205.11822}, 2022.

\bibitem[Krishna et~al.(2024)Krishna, Agarwal, and Lakkaraju]{krishna2024understandingeffectsiterativeprompting}
Satyapriya Krishna, Chirag Agarwal, and Himabindu Lakkaraju.
\newblock Understanding the effects of iterative prompting on truthfulness, 2024.
\newblock URL \url{https://arxiv.org/abs/2402.06625}.

\bibitem[Li et~al.(2024)Li, Zhang, Yu, Fu, and Ye]{li2024more}
Junyou Li, Qin Zhang, Yangbin Yu, Qiang Fu, and Deheng Ye.
\newblock More agents is all you need.
\newblock \emph{arXiv preprint arXiv:2402.05120}, 2024.

\bibitem[Liu et~al.(2024)Liu, Yao, Zhang, Yang, Liu, Tan, Choubey, Lan, Wu, Wang, et~al.]{liu2024agentlite}
Zhiwei Liu, Weiran Yao, Jianguo Zhang, Liangwei Yang, Zuxin Liu, Juntao Tan, Prafulla~K Choubey, Tian Lan, Jason Wu, Huan Wang, et~al.
\newblock Agentlite: A lightweight library for building and advancing task-oriented llm agent system.
\newblock \emph{arXiv preprint arXiv:2402.15538}, 2024.

\bibitem[Madaan et~al.(2024)Madaan, Tandon, Gupta, Hallinan, Gao, Wiegreffe, Alon, Dziri, Prabhumoye, Yang, et~al.]{madaan2024self}
Aman Madaan, Niket Tandon, Prakhar Gupta, Skyler Hallinan, Luyu Gao, Sarah Wiegreffe, Uri Alon, Nouha Dziri, Shrimai Prabhumoye, Yiming Yang, et~al.
\newblock Self-refine: Iterative refinement with self-feedback.
\newblock \emph{Advances in Neural Information Processing Systems}, 36, 2024.

\bibitem[OpenAI(2023)]{openai2023gpt4}
OpenAI.
\newblock Gpt-4 technical report.
\newblock \emph{OpenAI Technical Report}, 2023.
\newblock URL \url{https://cdn.openai.com/papers/gpt-4.pdf}.
\newblock Accessed: 2023-09-13.

\bibitem[OpenAI(2024)]{OpenAI2024eval}
OpenAI.
\newblock Simple evals.
\newblock \url{https://github.com/openai/simple-evals}, 2024.

\bibitem[{OpenAI}(2024)]{openai2024o1}
{OpenAI}.
\newblock Introducing openai o1-preview.
\newblock \url{https://openai.com/index/introducing-openai-o1-preview/}, September 2024.
\newblock URL \url{https://openai.com/index/introducing-openai-o1-preview/}.
\newblock Accessed: 2024-09-17.

\bibitem[Opsahl-Ong et~al.(2024)Opsahl-Ong, Ryan, Purtell, Broman, Potts, Zaharia, and Khattab]{opsahl2024optimizing}
Krista Opsahl-Ong, Michael~J Ryan, Josh Purtell, David Broman, Christopher Potts, Matei Zaharia, and Omar Khattab.
\newblock Optimizing instructions and demonstrations for multi-stage language model programs.
\newblock \emph{arXiv preprint arXiv:2406.11695}, 2024.

\bibitem[Petroni et~al.(2019)Petroni, Rockt{\"a}schel, Lewis, Bakhtin, Wu, Miller, and Riedel]{petroni2019language}
Fabio Petroni, Tim Rockt{\"a}schel, Patrick Lewis, Anton Bakhtin, Yuxiang Wu, Alexander~H Miller, and Sebastian Riedel.
\newblock Language models as knowledge bases?
\newblock \emph{arXiv preprint arXiv:1909.01066}, 2019.

\bibitem[Pydantic(2024)]{pydantic_2024}
Pydantic.
\newblock pydantic-2.9.2.
\newblock \url{https://pypi.org/project/pydantic}, 2024.
\newblock URL \url{https://pypi.org/project/pydantic}.
\newblock Data validation using Python type hints.

\bibitem[Rein et~al.(2023)Rein, Hou, Stickland, Petty, Pang, Dirani, Michael, and Bowman]{rein2023gpqa}
David Rein, Betty~Li Hou, Asa~Cooper Stickland, Jackson Petty, Richard~Yuanzhe Pang, Julien Dirani, Julian Michael, and Samuel~R Bowman.
\newblock Gpqa: A graduate-level google-proof q\&a benchmark.
\newblock \emph{arXiv preprint arXiv:2311.12022}, 2023.

\bibitem[Roberts et~al.(2020)Roberts, Raffel, and Shazeer]{roberts2020much}
Adam Roberts, Colin Raffel, and Noam Shazeer.
\newblock How much knowledge can you pack into the parameters of a language model?
\newblock \emph{arXiv preprint arXiv:2002.08910}, 2020.

\bibitem[Sahoo et~al.(2024)Sahoo, Singh, Saha, Jain, Mondal, and Chadha]{sahoo2024systematic}
Pranab Sahoo, Ayush~Kumar Singh, Sriparna Saha, Vinija Jain, Samrat Mondal, and Aman Chadha.
\newblock A systematic survey of prompt engineering in large language models: Techniques and applications.
\newblock \emph{arXiv preprint arXiv:2402.07927}, 2024.

\bibitem[Sloman(1996)]{sloman1996empirical}
Steven~A Sloman.
\newblock The empirical case for two systems of reasoning.
\newblock \emph{Psychological bulletin}, 119\penalty0 (1):\penalty0 3, 1996.

\bibitem[Talmor et~al.(2020)Talmor, Elazar, Goldberg, and Berant]{talmor2020olmpics}
Alon Talmor, Yanai Elazar, Yoav Goldberg, and Jonathan Berant.
\newblock olmpics-on what language model pre-training captures.
\newblock \emph{Transactions of the Association for Computational Linguistics}, 8:\penalty0 743--758, 2020.

\bibitem[Team et~al.(2023)Team, Anil, Borgeaud, Wu, Alayrac, Yu, Soricut, Schalkwyk, Dai, Hauth, et~al.]{team2023gemini}
Gemini Team, Rohan Anil, Sebastian Borgeaud, Yonghui Wu, Jean-Baptiste Alayrac, Jiahui Yu, Radu Soricut, Johan Schalkwyk, Andrew~M Dai, Anja Hauth, et~al.
\newblock Gemini: a family of highly capable multimodal models.
\newblock \emph{arXiv preprint arXiv:2312.11805}, 2023.

\bibitem[Tonmoy et~al.(2024)Tonmoy, Zaman, Jain, Rani, Rawte, Chadha, and Das]{tonmoy2024comprehensive}
SM~Tonmoy, SM~Zaman, Vinija Jain, Anku Rani, Vipula Rawte, Aman Chadha, and Amitava Das.
\newblock A comprehensive survey of hallucination mitigation techniques in large language models.
\newblock \emph{arXiv preprint arXiv:2401.01313}, 2024.

\bibitem[Von~Bertalanffy(1950)]{von1950theory}
Ludwig Von~Bertalanffy.
\newblock The theory of open systems in physics and biology.
\newblock \emph{Science}, 111\penalty0 (2872):\penalty0 23--29, 1950.

\bibitem[Wang et~al.(2024{\natexlab{a}})Wang, Lu, Santacroce, Gong, Zhang, et~al.]{wang2024adapting}
Kuan Wang, Yadong Lu, Michael Santacroce, Yeyun Gong, Chao Zhang, et~al.
\newblock Adapting llm agents with universal feedback in communication.
\newblock In \emph{ICML 2024 Workshop on Foundation Models in the Wild}, 2024{\natexlab{a}}.

\bibitem[Wang et~al.(2022)Wang, Wei, Schuurmans, Le, Chi, Narang, Chowdhery, and Zhou]{wang2022self}
Xuezhi Wang, Jason Wei, Dale Schuurmans, Quoc Le, Ed~Chi, Sharan Narang, Aakanksha Chowdhery, and Denny Zhou.
\newblock Self-consistency improves chain of thought reasoning in language models.
\newblock \emph{arXiv preprint arXiv:2203.11171}, 2022.

\bibitem[Wang et~al.(2024{\natexlab{b}})Wang, Lipka, Rossi, Siu, Zhang, and Derr]{wang2024knowledge}
Yu~Wang, Nedim Lipka, Ryan~A Rossi, Alexa Siu, Ruiyi Zhang, and Tyler Derr.
\newblock Knowledge graph prompting for multi-document question answering.
\newblock In \emph{Proceedings of the AAAI Conference on Artificial Intelligence}, volume~38, pages 19206--19214, 2024{\natexlab{b}}.

\bibitem[Wei et~al.(2022)Wei, Wang, Schuurmans, Bosma, Xia, Chi, Le, Zhou, et~al.]{wei2022chain}
Jason Wei, Xuezhi Wang, Dale Schuurmans, Maarten Bosma, Fei Xia, Ed~Chi, Quoc~V Le, Denny Zhou, et~al.
\newblock Chain-of-thought prompting elicits reasoning in large language models.
\newblock \emph{Advances in neural information processing systems}, 35:\penalty0 24824--24837, 2022.

\bibitem[Weng et~al.(2022)Weng, Zhu, Xia, Li, He, Liu, Sun, Liu, and Zhao]{weng2022large}
Yixuan Weng, Minjun Zhu, Fei Xia, Bin Li, Shizhu He, Shengping Liu, Bin Sun, Kang Liu, and Jun Zhao.
\newblock Large language models are better reasoners with self-verification.
\newblock \emph{arXiv preprint arXiv:2212.09561}, 2022.

\bibitem[Wu et~al.(2022)Wu, Xiao, Sun, Zhang, Ma, and He]{wu2022survey}
Xingjiao Wu, Luwei Xiao, Yixuan Sun, Junhang Zhang, Tianlong Ma, and Liang He.
\newblock A survey of human-in-the-loop for machine learning.
\newblock \emph{Future Generation Computer Systems}, 135:\penalty0 364--381, 2022.

\bibitem[Yao et~al.(2024)Yao, Yu, Zhao, Shafran, Griffiths, Cao, and Narasimhan]{yao2024tree}
Shunyu Yao, Dian Yu, Jeffrey Zhao, Izhak Shafran, Tom Griffiths, Yuan Cao, and Karthik Narasimhan.
\newblock Tree of thoughts: Deliberate problem solving with large language models.
\newblock \emph{Advances in Neural Information Processing Systems}, 36, 2024.

\end{thebibliography}
\newpage
%%% Appendix Section
\appendix
\section{Appendix}
\label{appendix:experimental_details}

\subsection{Examples}
In this section, we provide an example to demonstrate how AIoT (Autonomous Iteration of Thought) works in practice. This example highlights the unique characteristics of AIoT, illustrating how the method improves reasoning, adaptability, and response accuracy.

\subsubsection{Example of AIoT}

This example highlights AIoT's strength in efficiently navigating a complex GPQA example.
\begin{iotenv}{}
A textile dye containing an extensively conjugated pi-electrons emits light with energy of 2.3393 eV. What color of light is absorbed by the organic compound? Pick an answer from the following options: \\
A. Red \\
B. Yellow \\
C. Blue \\
D. Violet
\end{iotenv}

% Iteration 1
\addida{
\textbf{Iteration 1:} \\
Let's start by analyzing the problem. The energy of the light emitted by the dye is given as 2.3393 eV. To determine the color of light absorbed by the dye, we need to recall that the color emitted is complementary to the color that is absorbed. We need to convert the energy of the emitted light into the wavelength to find out the corresponding color. Using the formula $E = \frac{hc}{\lambda}$, where $h$ is Planck's constant and $c$ is the speed of light, we can calculate the wavelength. Can you recall the values of Planck's constant and the speed of light to perform this calculation?
}

\addllma{
Let's indeed gather the necessary constants for the calculations: \\
- Planck's constant ($h$) is approximately $6.626 \times 10^{-34} \, \text{Js}$ \\
- The speed of light ($c$) is about $3.00 \times 10^8 \, \text{m/s}$ \\

We can convert the energy of the emitted light (2.3393 eV) to joules first, using the conversion factor $1 \, \text{eV} = 1.602 \times 10^{-19} \, \text{J}$: \\
\[
E = 2.3393 \, \text{eV} \times 1.602 \times 10^{-19} \, \text{J/eV} = 3.75 \times 10^{-19} \, \text{J}
\] \\
Now we can use the relationship $E = \frac{hc}{\lambda}$ to find the wavelength $(\lambda)$: \\
\[
\lambda = \frac{hc}{E}
\] \\
Substituting the known values: \\
\[
\lambda = \frac{(6.626 \times 10^{-34} \, \text{Js}) (3.00 \times 10^8 \, \text{m/s})}{3.75 \times 10^{-19} \, \text{J}}
\] \\
Now let’s calculate that expression.
}

% Iteration 2
\addida{
\textbf{Iteration 2:} \\
Let's perform the calculations step by step. First, substitute the values of Planck's constant and the speed of light into the wavelength formula. Once you have calculated the wavelength, recall that color is associated with specific wavelengths in the electromagnetic spectrum. What are the ranges of wavelengths for red, yellow, blue, and violet light? After you calculate the wavelength, identify which color corresponds to that range and determine the complementary color that would be absorbed by the dye.
}

\addllma{
In order to find the wavelength associated with the emitted energy of 2.3393 eV, let's first calculate the value of $\lambda$ using the formula: \\
\[
\lambda = \frac{h c}{E}
\] \\
Substituting in the known values: \\
- Planck's constant, $h = 6.626 \times 10^{-34} \, \text{Js}$ \\
- Speed of light, $c = 3.00 \times 10^8 \, \text{m/s}$ \\
- Energy, $E = 3.75 \times 10^{-19} \, \text{J}$ \\

Calculating it step by step: \\
1. First calculate the numerator: \\
   \[
   h c = (6.626 \times 10^{-34} \, \text{Js}) (3.00 \times 10^8 \, \text{m/s}) = 1.9878 \times 10^{-25} \, \text{J m}
   \] \\
2. Now divide by the energy: \\
   \[
   \lambda = \frac{1.9878 \times 10^{-25} \, \text{J m}}{3.75 \times 10^{-19} \, \text{J}}
   \] \\
   \[
   \lambda \approx 5.2968 \times 10^{-7} \, \text{m} = 529.68 \times 10^{-9} \, \text{m} = 529.68 \, \text{nm}
   \] \\
Now that we have calculated the wavelength, we can relate it to its corresponding color: \\
- The visible light spectrum generally ranges from roughly 380 nm (violet) to 750 nm (red). \\
- The color associated with $529.68 \, \text{nm}$ falls within the green range of visible light. \\

Based on the concept of complementary colors, we see that green has a complementary color of red, meaning that if the dye emits green light, it would absorb red light. \\
Next, let's summarize our findings: \\
- Emitted light: $529.68 \, \text{nm}$ (green) \\
- Absorbed light: red \\

Since red is the only option on the list that matches our analysis, I am now ready to conclude this iteration.
}

% Iteration 3
\addida{
\textbf{Iteration 3:} \\
You have correctly calculated the emitted wavelength of light (529.68 nm), which corresponds to the green portion of the spectrum. You've also established that the complementary color of green is red, which means if the dye emits green light, it absorbs red light. However, it’s important to clearly differentiate the roles of emitted and absorbed light in the context of pi-conjugated systems to ensure there is no confusion in conclusions drawn. Also, let's briefly verify the wavelength ranges for more clarity.
}

\addllma{
You bring up a crucial point regarding the roles of emitted and absorbed light, particularly in the context of pi-conjugated systems. Let's review the specific ranges of wavelengths for the colors mentioned: \\
- **Red**: ~620 to 750 nm \\
- **Yellow**: ~570 to 590 nm \\
- **Blue**: ~450 to 495 nm \\
- **Violet**: ~380 to 450 nm \\

Given our calculated wavelength of approximately 529.68 nm, we acknowledged earlier that this corresponds to the green region of the spectrum, which indeed aligns well with the understanding that it is complementary to red light in this case. \\

To verify one more time: \\
- The emitted light being in the green spectrum suggests that it would not absorb light in that range. Instead, it indicates that within a complementary color framework, if our emission is at 529.68 nm, the absorption would typically occur at a wavelength corresponding to the red light range. \\

Thus, what we could summarize so far is: \\
- The dye emits green light (529.68 nm) \\
- The dye absorbs light that is complementary to green, which is red. \\

Since we have the absorption specifically related to red light, and knowing that red is an acceptable answer option available, this aligns with the logical deductions we've made. I think we are edging closer to a definitive conclusion! \\

Is there anything else you would like to explore further? Or should we finalize our answer regarding the absorbed light being red?
}

% Iteration 4
\addida{
\textbf{Iteration 4:} \\
You've gathered a robust understanding of the emission at 529.68 nm, which corresponds to green light. Remember that the dye absorbs light that is the complementary color to what it emits. Thus, while we've confirmed the complementary relationship of green to red, it’s important to consider if this is sufficient for the conclusion or if you want to ensure thorough reasoning by re-evaluating the answer choices in the context of shorter wavelength absorption. What is your final answer considering the options provided?
}

\addllma{
To summarize our progress thus far: \\
1. We calculated the emitted energy of 2.3393 eV translated to an approximate wavelength of 529.68 nm, which falls within the green region of the visible spectrum. \\
2. We identified that the complementary color of emitted light (green) is red. Therefore, if the textile dye emits green light, it absorbs red light. \\
3. The provided options are: \\
   - A. Red \\
   - B. Yellow \\
   - C. Blue \\
   - D. Violet \\

Given this analysis, the only option that fits our conclusion that the dye absorbs red light is option A. \\

However, it is essential to ensure our understanding involves comprehensive reasoning; we considered various potential nuances regarding other colors that could be absorbed based on the broader electromagnetic spectrum and the nature of pi-conjugated systems. As we reflect on the absence of additional factors influencing absorption, our primary focus remains the complementary nature of absorption versus emission. \\

To firmly establish clarity: \\
- If the dye is emitting green light due to its molecular structure with extended pi-electron conjugation, it's plausible that the absorption spectrum of this compound overlaps significantly with red light. \\
- Furthermore, it's prudent to assure that the chosen answer remains consistent with the well-understood principles in color theory and the relationship between absorbed and emitted light. \\

Thus we appear to have reached solid conclusions with our reasoning up to this point. Would you like to explore any specific case studies or examples regarding similar spectral behaviors, or feel satisfied to proceed towards concluding the correct answer based on your questions and considerations?
}

% Iteration 5
\addida{
\textbf{Iteration 5:} \\
We have thoroughly analyzed the relationships between emitted and absorbed light concerning the organic dye's properties due to its extensively conjugated pi-electrons. We accurately calculated the emitted wavelength corresponding to the energy given and identified it as green (529.68 nm). It is established that the dye should absorb the complementary color of the emitted light, which is red. The options provided include red, yellow, blue, and violet. Since red is the only color that matches the conclusion drawn from our reasoning about absorption, we can affirm that our answer aligns with the correct scientific principles and correspondence. Thus, we can finalize our answer as red. Would you like to summarize any additional points or confirm if you are content with this outcome?
}

\addllma{
Based on our analysis of the problem, we have reached a comprehensive understanding of how to approach the relationship between the absorbed and emitted light of a substance with extensively conjugated pi-electrons. Here's a final breakdown of our findings: \\
1. **Emission and Calculated Wavelength**: We've established that the energy of the emitted light is 2.3393 eV, leading us to calculate a corresponding wavelength of approximately 529.68 nm, which lies within the green region of the visible spectrum. \\
2. **Understanding Complementary Colors**: We recalled that each color has a complementary counterpart in the context of absorption and emission. Since our calculated wavelength (green) requires understanding that red is the complementary color to green based on color theory principles. \\
3. **Re-evaluation of Answer Choices**: \\
   - A. **Red** - complement of green. \\
   - B. **Yellow** - not complementary to green. \\
   - C. **Blue** - also not complementary. \\
   - D. **Violet** - not complementary. \\

Given that red is indeed complementary to the identified green emission, it supports the conclusion that the organic compound absorbs red light. \\

Thus, we can confidently finalize: \\
The absorbed color of light by the organic compound is **A. Red**. Would you like to confirm this conclusion or discuss anything further?
}

% Final Answer Box
\finalanswer{
The textile dye emits light corresponding to a wavelength of approximately 529.68 nm, which is in the green portion of the visible spectrum. The color that is complementary to green is red. Therefore, the dye absorbs red light. \\
\textbf{Answer: A (Red)}
}

\end{document}